\pdfoutput=1

\documentclass[11pt, dvipsnames]{article}

\usepackage[]{acl}

\usepackage{times}
\usepackage{latexsym}
\usepackage{float}
\usepackage{graphicx}
\usepackage{xcolor}
\usepackage{soul}
\usepackage{colortbl}
\usepackage{color}
\usepackage{caption}   
\usepackage{subcaption}  % 用于子图
\usepackage{tabularx}
\usepackage{dingbat}
\usepackage{pifont}
\usepackage{amsmath}
\usepackage{amssymb}
\usepackage{makecell}
\usepackage{arydshln}
\usepackage{stfloats}
\usepackage[T1]{fontenc}
\usepackage{titlesec}
\usepackage{enumitem} % 在导言区引入
\usepackage[utf8]{inputenc}
\usepackage{CJKutf8}
\usepackage{multirow}
\usepackage{array}
\usepackage[utf8]{inputenc}
\usepackage{microtype}
\usepackage{booktabs}
\usepackage{xcolor}
\usepackage{fontawesome5}
\usepackage{caption}
\usepackage{amsfonts}
\usepackage{inconsolata}
\usepackage{arydshln}
\usepackage{CJKutf8}
\usepackage{fancyvrb}
\usepackage{listings}
\usepackage{hyperref}
\usepackage[most]{tcolorbox}
\definecolor{ReActColor}{HTML}{6495ed}
\definecolor{ExitColor}{HTML}{fdc1bd}

\newcommand{\cmark}{{\textbf{\textcolor[rgb]{0.1, 0.5, 0.1}{\ding{51}}}}}
\newcommand{\xmark}{{\textbf{\color{red}{\ding{55}}}}}

\definecolor{myblue}{rgb}{0.82, 0.94, 0.75}

\title{Runaway is Ashamed, But Helpful: On the Early-Exit Behavior of Large Language Model-based Agents in Embodied Environments}

\usepackage{pifont}

\author{
\textbf{
Qingyu~Lu$^{\diamondsuit\text{\ding{95}}}$\thanks{~~Equal contribution.}, 
Liang Ding$^{\heartsuit}$\footnotemark[1], 
Siyi Cao$^{\diamondsuit}$, 
Xuebo Liu$^{\text{\ding{100}}}$, 
Kanjian Zhang$^{\diamondsuit\spadesuit}$\thanks{~~Corresponding Author.}
} \\
\textbf{Jinxia Zhang$^{\diamondsuit}$, 
Dacheng Tao$^{\text{\ding{95}}}$} \\
\fontsize{10pt}{12pt}\selectfont $^{\diamondsuit}$Southeast University \
$^{\heartsuit}$The University of Sydney \
$^{\spadesuit}$Southeast University Shenzhen Research Institute \\
\fontsize{10pt}{12pt}\selectfont $^{\text{\ding{100}}}$Institute of Computing and Intelligence, Harbin Institute of Technology, Shenzhen, China \\
\fontsize{10pt}{12pt}\selectfont $^{\text{\ding{95}}}$College of Computing and Data Science at Nanyang Technological University, Singapore 639798 \\
\raisebox{-0.6ex}{\includegraphics[scale=0.032]{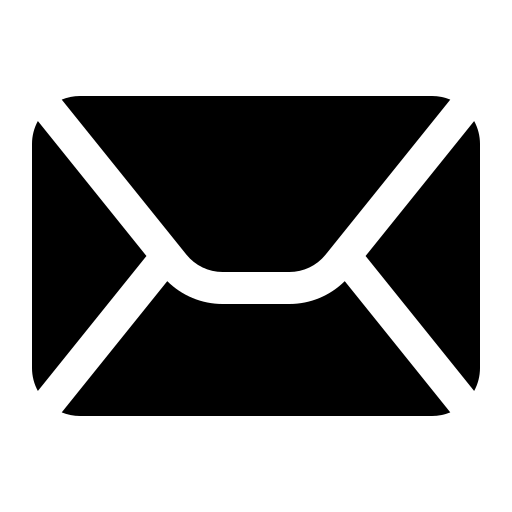}} \fontsize{9pt}{9pt}\selectfont \texttt{\{luqingyu,siyic,kjzhang,jinxiazhang\}@seu.edu.cn}, \texttt{liangding.liam@gmail.com}, \\
\fontsize{9pt}{9pt}\selectfont \texttt{liuxuebo@hit.edu.cn}, \texttt{dacheng.tao@gmail.com}\\
\raisebox{-0.35ex}{\includegraphics[scale=0.03]{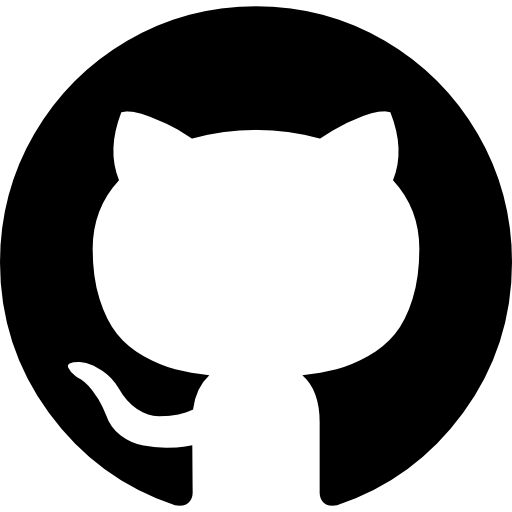}} \fontsize{9pt}{10pt}\selectfont \url{https://github.com/Coldmist-Lu/AgentExit}
}

\begin{document}
\maketitle
\begin{abstract}

Agents powered by large language models (LLMs) have demonstrated strong planning and decision-making capabilities in complex embodied environments. However, such agents often suffer from inefficiencies in multi-turn interactions, frequently trapped in repetitive loops or issuing ineffective commands, leading to redundant computational overhead. Instead of relying solely on learning from trajectories, we take a first step toward exploring the early-exit behavior for LLM-based agents. We propose two complementary approaches, \ding{182} an \textbf{intrinsic} method that injects exit instructions during generation, and \ding{183} an \textbf{extrinsic} method that verifies task completion to determine when to halt an agent’s trial. To evaluate early-exit mechanisms, we introduce two metrics: one measures the reduction of \textbf{redundant steps} as a positive effect, and the other evaluates \textbf{progress degradation} as a negative effect. Experiments with 4 different LLMs across 5 embodied environments show significant efficiency improvements, with only minor drops in agent performance. We also validate a practical strategy where a stronger agent assists after an early-exit agent, achieving better performance with the same total steps. We will release our code to support further research.

\end{abstract}

\section{Introduction}

\begin{figure}[t]
\centering
\includegraphics[scale=0.75]{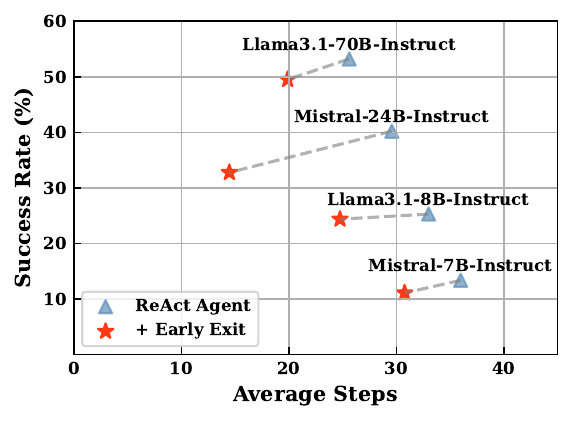}
\caption{\textbf{Early-exit behavior of different LLM-based agents} in embodied environments. While early termination slightly reduces the success rate, it significantly decreases the average number of interaction steps, indicating improved efficiency.}
\label{fig:avg_res}
\end{figure}

\begin{figure*}[t]
\centering
\includegraphics[scale=0.56]{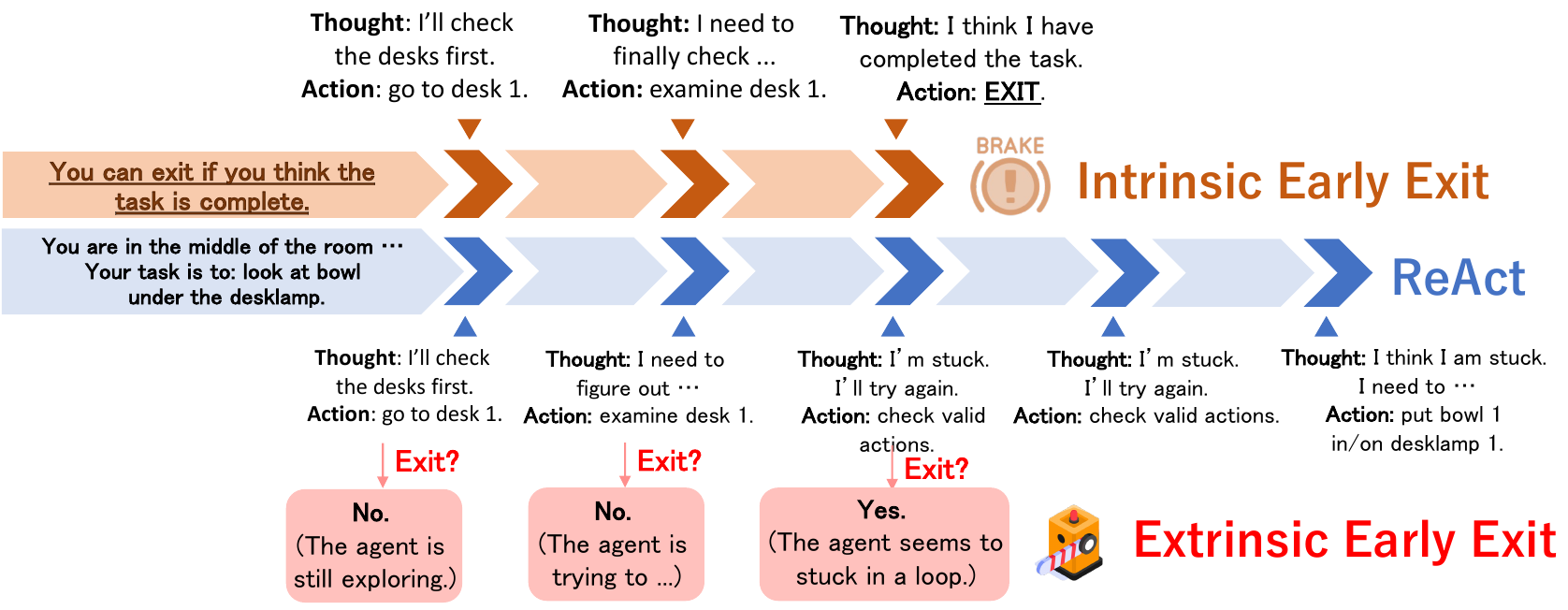}
\caption{\textbf{A comparative overview of our proposed Intrinsic and Extrinsic Early Exit} with ReAct Agent. The intrinsic approach injects an exit instruction to guide the agent to self-terminate, while the extrinsic approach uses a verification module to determine whether to exit based on the current state.}
\label{fig:overview}
\end{figure*}

Large Language Models (LLMs, \citealp{achiam2023gpt}) have shifted the paradigm from merely responding to user inputs to tackling more complex tasks within interactive environments such as household settings \citep{shridharalfworld}, virtual worlds \citep{park2023genagents}, and games \citep{hu2024survey}. LLM-based agents serve as intelligent controllers, capable of perceiving environments, executing actions, and adapting through feedback \citep{wang2024survey, luo2025large}. Previous studies show that structured workflows—such as reasoning before acting \citep{yao2023react}, predicting future states \citep{fu-etal-2025-preact}, and learning from high-quality trajectories \citep{chen-etal-2024-agent, song-etal-2024-trial}—can improve performance within a single trial. When agents do fail, post-hoc approaches such as Reflexion \citep{shinn2023reflexion}, AutoPlan \citep{ouyang-li-2023-autoplan}, and ExpeL \citep{zhao2024expel} enable them to learn from failures and replan more effective solutions in subsequent trials.

However, a key limitation of LLM-based agents remains underexplored: \textit{they often fail to recognize when a goal is too difficult or when they are stuck}. Prior work shows that agents may repeat the same errors in unproductive loops without meaningful actions or self-correction \citep{fuagentrefine}, leading to unnecessary computational overhead. This issue becomes even more critical in real-world settings, where repeated mistakes by embodied agents can waste energy, cause wear-and-tear, or even damage physical objects in the environment. Therefore, incorporating built-in self-awareness mechanisms can help agents detect when progress has stalled, enabling early self-reflection and adjustment.

To this end, we take the first step by investigating the \textit{early-exit} behavior of LLM-based agents. As shown in Figure~\ref{fig:overview}, we propose two complementary strategies: \ding{182} \textbf{Intrinsic Early Exit}, which injects exit instructions directly into the agent's prompts to encourage self-recognition of when to halt; and \ding{183} \textbf{Extrinsic Early Exit}, which introduces an external verification module that monitors the interaction status and outputs a binary (YES/NO) decision to control whether the agent should continue.

In addition to using success rate and progress rate \citep{chang2024agentboard} to evaluate agent performance, we propose two new metrics to assess the impact of the early-exit mechanism. \textit{Redundant Steps} quantifies the positive effect by measuring reductions in unnecessary interactions, while \textit{Progress Degradation} captures the potential negative impact, indicating cases where exit early may interrupt or reverse meaningful progress. 

We conduct experiments on 5 datasets spanning over 400 environments and find that the early-exit mechanism significantly improves efficiency, with only a minor drop in task success and progress rates, as shown in Figure~\ref{fig:avg_res}. We also propose a practical use of early-exit behavior: Once the agent exits early, a stronger agent reflects on the state and continues exploration, achieving improved performance within the same total steps.

Our contributions are three-fold:

\begin{itemize}
    \item We present the first investigation into early-exit behavior in LLM-based agents, proposing two strategies that enable agents to develop self-awareness and terminate execution without external intervention.
    \item We introduce two complementary metrics to evaluate the effectiveness of early exit. These metrics can serve as standardized tools for assessing agent behavior and guiding the selection of optimal exiting strategies.
    \item Our proposed methods generalize across various LLM-based agents and task settings. We further demonstrate the practical value of our approach by introducing post-trial strategies that leverage stronger agents to enhance overall performance.
\end{itemize}

This study is an initial step toward exploring early-exit behavior in LLM-based agents. Our approach encourages agents to make efficient decisions, avoid unnecessary interactions, and achieve a trade-off between efficiency and task performance.

\section{Approach}

\subsection{Task Formulation}

\paragraph{Embodied Environments} In embodied environments, an agent interacts with the world through actions and receives feedback from the environment. This interaction can be modeled as a special case of a Partially Observable Markov Decision Process (POMDP), defined by an instruction space \( U \), state space \( S \), action space \( A \), observation space \( O \), and a transition function \( T: S \times A \to S \).

\paragraph{LLM-based Agents} 
In this work, we focus on text-based environments, where the instruction, action, and observation spaces are all expressed in natural language. The agent is provided with an instruction \( u \), which includes a description of the task and environment, as well as the goal to be achieved. At each time step \( t \), the agent, guided by a policy \( \pi_\theta \) (typically an LLM with parameters \( \theta \)), must decide on the next action \( a_t \) based on the trajectory history \( e_t \). This decision-making process is formalized as:
\begin{equation}
    a_t \sim \pi_\theta(\cdot \mid e_t, u),
\end{equation}
where \( e_t = (a_1, o_1, \ldots, a_{t-1}, o_{t-1}) \) denotes the full trajectory up to time \( t \), including previous actions and observations. In this way, the agent continually explores the environment, using feedback from observations to inform its next actions, until the task is completed or a predefined maximum number of steps is reached.

\subsection{Dynamic Early Exit}

We propose two simple but effective early-exit strategies, \textit{Intrinsic Early Exit} and \textit{Extrinsic Early Exit}, that enable the agent to terminate its interaction when appropriate.

\paragraph{Intrinsic Early Exit} 
This strategy modifies the behavior of LLM agent by appending a natural language prompt that allows it to terminate the interaction with the environment when deemed necessary. The exit instruction can be formulated as:
\begin{equation}
    u_{\text{intrinsic}} = \text{concat}(u, u_{\text{exit}}).
\end{equation}
In this way, the LLM may develop an intention to termiate based on the additional instruction \( u_{\text{exit}} \), leading to different actions and trajectories. As shown in Figure~\ref{fig:overview}, the agent is prompted with an instruction to exit once the task is complete. After examining the relevant objects, the agent generates an "EXIT" action to terminate the interaction.

\paragraph{Extrinsic Early Exit}
This strategy introduces a verification module \( v_\theta \), which shares the same LLM backbone. The verification module operates after each action and observation, evaluating whether the agent should continue the task. It outputs a binary decision: "YES" to exit or "NO" to continue execution. Specifically, it functions as follows:
\begin{align}
    &u_{\text{extrinsic}} = \text{concat}(u, u_{\text{exit}}), \\
    &v_\theta(\cdot \mid e_t, u_{\text{extrinsic}}) \in \{0, 1\}.
\end{align}
The agent is verified periodically every \( k \) steps. In our experiments, we set \( k=1 \)\footnote{We set \(k=1\) to enable timely detection in our experiments. In practice, larger values (e.g., \(k=2\)–\(5\)) can be used to reduce computational overhead.}. As shown in Figure~\ref{fig:overview}, the verification module detects that the agent is stuck and triggers an early exit, effectively avoiding further repetitive steps.

\subsection{Evaluation}

\begin{figure*}[t]
  \centering
  \begin{subfigure}[b]{0.47\textwidth}
    \includegraphics[width=\textwidth]{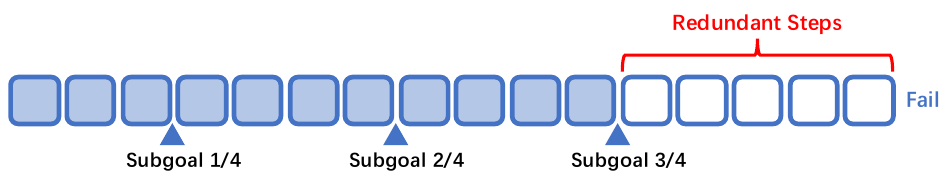}
    \caption*{(a) Redundant Steps}
  \end{subfigure}
  \hspace{0.02\textwidth}
  \begin{subfigure}[b]{0.47\textwidth}
    \includegraphics[width=\textwidth]{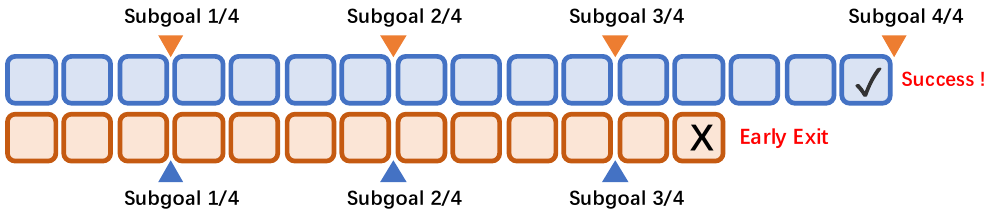}
    \caption*{(b) Progress Degradation}
  \end{subfigure}
  \caption{\textbf{An overview of the proposed metrics}. \textit{Redundant Steps} measures the number of redundant steps. \textit{Progress Degradation} measures task progress loss via reduced subgoal completion.}
  \label{fig:newmetrics}
\end{figure*}

Typically, the performance of agents in embodied environments is evaluated using Success Rate and Progress Rate. To intuitively demonstrate the behavior of the early-exit mechanism on LLM-based agents, we propose two complementary metrics that capture both its positive and negative effects. These metrics are defined as follows:

\paragraph{Success Rate (SR)}
The environment is marked as successful if the agent completes the given task, typically when it reaches a predefined latent state that signifies task completion. A higher success rate indicates that the agent is more effective at solving environments under the same task.

\paragraph{Progress Rate (PR)}  
Progress Rate, proposed by \citet{chang2024agentboard}, quantifies the extent to which an agent advances toward the task goal, making it particularly valuable for evaluating incremental improvements. In embodied environments, the task goal is decomposed into a sequence of subgoals \( G = [g_1, \cdots, g_K] \), where each subgoal contributes progressively to task completion. At each time step \( t \), the progress is defined as:
\begin{equation}
    r_t = \max_{i, 0 \leq i \leq t} \left( \frac{1}{K} \sum_{k=1}^K f(s_i, g_k) \right),
\end{equation}
where \( f(s_i, g_k) \in \{0, 1\} \) is a binary indicator function that evaluates whether the agent state \( s_i \) satisfies subgoal \( g_k \), typically determined via regular-expression-based matching. PR offers a more fine-grained and informative evaluation of agent behavior than binary success metrics alone.

\paragraph{New Metric 1: Redundant Steps (RS)}  
The primary purpose of introducing the early-exit mechanism is to reduce redundant steps in the agent's interaction with the environment. As illustrated in Figure~\ref{fig:newmetrics}(a), after completing subgoal 3 out of 4, the agent continues exploring unnecessarily for 5 additional steps before ultimately failing. Early-exit can mitigate this issue while maintaining the same level of progress. Let \( n_{\text{total}} \) denote the total number of steps in the trajectory, and \( n_{\text{subgoal}} \) be the index of the last step that achieves a new subgoal. The \textit{Redundant Steps} is defined as:
\begin{equation}
    \text{RS} = n_{\text{total}} - {n_{\text{subgoal}}}.
\end{equation}
For trivial cases, \(\text{RS} = n_{\text{total}}\) if the agent fails to complete any subgoal (i.e., \( \text{PR} = 0 \)). if the agent successfully completes the entire task, \( \text{RS} = 0 \), meaning that all steps are considered useful.

\paragraph{New Metric 2: Progress Degradation (PD)}  
The agent may also negatively impact agent performance by prematurely terminating trajectories that might have led to further progress. This can suppress the agent’s potential, causing missed subgoals or converting potentially successful trials into failures. To quantify this loss, we define \textit{Progress Degradation} as:
\begin{equation}
    \text{PD} = \max(\text{PR}_{\text{ref}} - \text{PR}_{\text{exit}}, 0),
\end{equation}
where \( \text{PR}_{\text{ref}} \) denotes the progress rate without exit, while \( \text{PR}_{\text{exit}} \) is the progress rate when early-exit is applied\footnote{Progress degradation is only meaningful when compared against a reference baseline.}. As shown in Figure~\ref{fig:newmetrics}(b), the agent exits 3 steps early, leaving an otherwise successful environment unfinished with only 75\% progress, resulting in a 25\% loss in progress. \textit{Progress Degradation} ranges from 0 (no degradation) to \( \text{PR}_{\text{ref}} \) (complete loss of progress). A higher PD indicates greater performance loss. In the trivial case, \( \text{PD} = 0 \) implies no degradation, while \( \text{PD} = \text{PR}_{\text{ref}} \) indicates complete progress failure (e.g., all environments terminate at the first step).

\section{Experimental Setup}

\paragraph{Datasets} We evaluate our methods across 3 embodied environments and 2 gaming environments. For embodied environments, \textbf{AlfWorld} \cite{shridharalfworld} includes 134 household tasks that require agents to explore their surroundings and complete instructions such as “Look at bowl under the desklamp.” \textbf{ScienceWorld} \citep{wang-etal-2022-scienceworld} simulates a total of 90 scientific experiments in an interactive setting, such as “measure the melting point.” \textbf{BabyAI} \cite{chevalierbabyai} is a 20x20 grid-based environment where agents must navigate and interact with objects to accomplish 112 defined goals. We also consider two gaming environments. \textbf{Jericho} \citep{hausknecht2020interactive} comprises 20 text-based fictional worlds, which we adapt using the setup from \citet{chang2024agentboard} to be completed within 15 subgoals. \textbf{PDDL} represents a suite of strategic planning tasks defined in the Planning Domain Definition Language \citep{vallati20152014}. Following \citet{chang2024agentboard}, we include four distinct games, namely, 60 unique environments for evaluation.

\paragraph{LLMs} To ensure reproducibility, we evaluate four open-source large language models with varying parameter sizes. From the LLaMA 3.1 series\footnote{\url{https://huggingface.co/meta-llama}} \citep{grattafiori2024llama}, developed by Meta, we use two instruction-tuned models: the 8B version (\textit{Llama3.1-8B-Instruct}) and the 70B version (\textit{Llama3.1-70B-Instruct}), with the latter quantized using 4-bit AWQ \citep{lin2024awq} for efficient inference. In addition, we test two models from the Mistral family\footnote{\url{https://huggingface.co/mistralai}} \citep{jiang2024mixtral}: \textit{Mistral-7B-Instruct} (v0.3) and \textit{Mistral-24B-Instruct} (Mistral-Small-Instruct-2409).

\paragraph{Prompts} We adopt ReAct-style \citep{yao2023react} prompting to enable LLM-based agents to interact effectively with the environment. Following \citet{song-etal-2024-trial}, we format the interaction prompt as a multi-turn conversation, including an in-context example for each task. For early-exit instructions, we explore prompt variants with varying strictness levels (see in Appendix~\ref{appendix:promptvariant}), aligning the strategy with specific LLMs.

\begin{table*}[t]
\centering
\small
\setlength{\tabcolsep}{4pt}
\begin{tabular}{ccccccccccccccccc}
\toprule[0.5mm]
\multicolumn{2}{c}{\textbf{Setting}} & \multicolumn{5}{c}{\textbf{ALFWorld}} & \multicolumn{5}{c}{\textbf{BabyAI}} & \multicolumn{5}{c}{\textbf{ScienceWorld}} \\
\cmidrule(lr){1-2} \cmidrule(lr){3-7} \cmidrule(lr){8-12} \cmidrule(lr){13-17}
\textbf{Int.} & \textbf{Ext.} & SR$\uparrow$ & PR$\uparrow$ & RS$\downarrow$ & PD$\downarrow$ & Steps$\downarrow$ & SR$\uparrow$ & PR$\uparrow$ & RS$\downarrow$ & PD$\downarrow$ & Steps$\downarrow$ & SR$\uparrow$ & PR$\uparrow$ & RS$\downarrow$ & PD$\downarrow$ & Steps$\downarrow$  \\
\midrule

\multicolumn{16}{c}{\textit{Llama3.1-8B-Instruct}} \\\hdashline[3pt/3pt]\noalign{\vskip 0.5ex}
 - & - & 23.1 & 45.2 & 26.4 & - & 33.4 & 41.1 & 54.6 & 18.3 & - & 27.1 & 8.9 & 37.3 & 29.5 & - & 38.5 \\
\cmark & \xmark 
& \cellcolor{red!9!white} 14.2 
& \cellcolor{red!7!white} 38.3 
& \cellcolor{green!30!white} 11.0 
& \cellcolor{red!14!white} 14.1 
& \cellcolor{green!35!white} 15.9 
& \cellcolor{red!0!white} 41.1 
& \cellcolor{red!0!white} 54.3 
& \cellcolor{green!6!white} 15.2 
& \cellcolor{red!10!white} 10.5 
& \cellcolor{green!3!white} 25.6 
& \cellcolor{red!1!white} 7.8 
& \cellcolor{red!4!white} 32.6 
& \cellcolor{green!8!white} 25.5 
& \cellcolor{red!11!white} 11.1 
& \cellcolor{green!12!white} 32.3 \\

\xmark & \cmark 
& \cellcolor{red!2!white} 20.9 
& \cellcolor{red!7!white} 38.3 
& \cellcolor{green!40!white} 6.1 
& \cellcolor{red!16!white} 16.3 
& \cellcolor{green!48!white} 9.6 
& \cellcolor{red!25!white} 16.1 
& \cellcolor{red!29!white} 25.4 
& \cellcolor{green!27!white} 4.9 
& \cellcolor{red!30!white} 30.5 
& \cellcolor{green!41!white} 6.6 
& \cellcolor{red!1!white} 7.8 
& \cellcolor{red!8!white} 29.5 
& \cellcolor{green!44!white} 7.2 
& \cellcolor{red!14!white} 13.9 
& \cellcolor{green!56!white} 10.5 \\

\cmark & \cmark 
& \cellcolor{red!1!white} 21.6 
& \cellcolor{red!1!white} 44.4 
& \cellcolor{green!30!white} 11.3 
& \cellcolor{red!9!white} 9.1 
& \cellcolor{green!33!white} 16.8 
& \cellcolor{green!11!white} 46.4 
& \cellcolor{green!5!white} 57.3 
& \cellcolor{green!4!white} 16.3 
& \cellcolor{red!6!white} 6.6 
& \cellcolor{green!4!white} 25.1 
& \cellcolor{red!1!white} 7.8 
& \cellcolor{red!3!white} 34.5 
& \cellcolor{green!9!white} 25.1 
& \cellcolor{red!9!white} 9.4 
& \cellcolor{green!12!white} 32.3 \\\midrule

\multicolumn{16}{c}{\textit{Llama3.1-70B-Instruct}} \\\hdashline[3pt/3pt]
\noalign{\vskip 0.5ex}
- & - & 76.1 & 81.1 & 7.2 & - & 19.0 & 49.1 & 62.8 & 16.1 & - & 26.4 & 34.4 & 67.5 & 17.9 & - & 31.4 \\
\cmark & \xmark 
& \cellcolor{red!14.9!white}61.2 
& \cellcolor{red!13.7!white}67.4 
& \cellcolor{green!3!white}5.7 
& \cellcolor{red!17.4!white}17.4 
& \cellcolor{green!20.8!white}13.8 
& \cellcolor{red!12.5!white}36.6 
& \cellcolor{red!9.7!white}53.1 
& \cellcolor{green!7!white}12.6 
& \cellcolor{red!18.0!white}18.0 
& \cellcolor{green!32.8!white}18.2 
& \cellcolor{red!15.5!white}18.9 
& \cellcolor{red!7.7!white}59.8 
& \cellcolor{green!15!white}10.3 
& \cellcolor{red!12.3!white}12.3 
& \cellcolor{green!40.8!white}21.2 \\
\xmark & \cmark 
& \cellcolor{red!5.9!white}70.2 
& \cellcolor{red!1.8!white}79.3 
& \cellcolor{green!6.8!white}3.8 
& \cellcolor{red!10.3!white}8.7 
& \cellcolor{green!22.4!white}13.4 
& \cellcolor{red!7.1!white}42.0 
& \cellcolor{red!3.5!white}59.3 
& \cellcolor{green!17!white}7.8 
& \cellcolor{red!13.0!white}12.9 
& \cellcolor{green!52.4!white}13.3 
& \cellcolor{red!6.6!white}27.8 
& \cellcolor{red!4.1!white}63.6 
& \cellcolor{green!22!white}6.8 
& \cellcolor{red!9.0!white}9.0 
& \cellcolor{green!56.0!white}17.4 \\
\cmark & \cmark 
& \cellcolor{green!17.6!white}80.6 
& \cellcolor{green!11.6!white}84.0 
& \cellcolor{green!6!white}4.3 
& \cellcolor{red!13.2!white}5.8 
& \cellcolor{green!8.0!white}17.0 
& \cellcolor{red!8.9!white}40.2 
& \cellcolor{red!5.0!white}57.8 
& \cellcolor{green!6.6!white}12.8 
& \cellcolor{red!13.3!white}13.1 
& \cellcolor{green!26.0!white}19.9 
& \cellcolor{red!6.6!white}27.8 
& \cellcolor{red!2.9!white}64.6 
& \cellcolor{green!13!white}11.1 
& \cellcolor{red!5.8!white}10.6 
& \cellcolor{green!34.4!white}22.8 \\\midrule
\multicolumn{16}{c}{\textit{Mistral-7B-Instruct}} \\\hdashline[3pt/3pt]\noalign{\vskip 0.5ex}
 - & - & 20.9 & 40.4 & 27.8 & - & 34.8 & 17.0 & 21.7 & 32.8 & - & 34.0 & 2.2 & 16.6 & 37.4 & - & 39.2 \\
\cmark & \xmark 
& \cellcolor{red!6.0!white}14.9 
& \cellcolor{red!4.1!white}36.3 
& \cellcolor{green!18!white}18.7 
& \cellcolor{red!15.9!white}15.9 
& \cellcolor{green!43.6!white}23.9 
& \cellcolor{red!0.9!white}16.1 
& \cellcolor{green!16.8!white}25.9 
& \cellcolor{green!7!white}29.2 
& \cellcolor{red!5.6!white}5.6 
& \cellcolor{green!8.0!white}32.0 
& \cellcolor{red!0.0!white}2.2 
& \cellcolor{green!7.2!white}18.4 
& \cellcolor{green!6.8!white}34 
& \cellcolor{red!3.4!white}3.4 
& \cellcolor{green!11.6!white}36.3 \\
\xmark & \cmark 
& \cellcolor{red!9.7!white}11.2 
& \cellcolor{red!7.9!white}32.5 
& \cellcolor{green!13.6!white}21 
& \cellcolor{red!18.1!white}16.7 
& \cellcolor{green!39.6!white}24.9 
& \cellcolor{red!6.3!white}10.7 
& \cellcolor{red!3.0!white}18.2 
& \cellcolor{green!12!white}26.8 
& \cellcolor{red!12.8!white}12.0 
& \cellcolor{green!24.4!white}27.9 
& \cellcolor{red!1.1!white}1.1 
& \cellcolor{red!1.2!white}15.4 
& \cellcolor{green!44.0!white}15.6 
& \cellcolor{red!3.4!white}3.4 
& \cellcolor{green!88.0!white}17.2 \\
\cmark & \cmark 
& \cellcolor{red!3.7!white}17.2 
& \cellcolor{red!4.3!white}36.1 
& \cellcolor{green!1.2!white}27.2 
& \cellcolor{red!4.4!white}11.4 
& \cellcolor{green!8.4!white}32.7 
& \cellcolor{red!0.9!white}16.1 
& \cellcolor{green!2.8!white}22.4 
& \cellcolor{green!2.8!white}31.4 
& \cellcolor{red!0.1!white}4.9 
& \cellcolor{green!2.0!white}33.5 
& \cellcolor{red!0.0!white}2.2 
& \cellcolor{green!6.4!white}18.2 
& \cellcolor{green!5.2!white}34.8 
& \cellcolor{red!4.7!white}1.9 
& \cellcolor{green!4.8!white}38.0 \\\midrule
\multicolumn{16}{c}{\textit{Mistral-24B-Instruct}} \\\hdashline[3pt/3pt]\noalign{\vskip 0.5ex}
 - & - & 58.2 & 71.6 & 11.7 & - & 25.9 & 49.1 & 60.9 & 17.1 & - & 25.5 & 15.6 & 42.5 & 26.4 & - & 36.9 \\
\cmark & \xmark 
& \cellcolor{red!26.9!white}31.3 
& \cellcolor{red!20!white}51.7 
& \cellcolor{green!3.2!white}10.1 
& \cellcolor{red!26.0!white}26.0 
& \cellcolor{green!36.0!white}17.4 
& \cellcolor{red!9.0!white}40.2 
& \cellcolor{red!9.5!white}51.5 
& \cellcolor{green!4.0!white}20.4 
& \cellcolor{red!6.0!white}19.4 
& \cellcolor{green!6.0!white}27.0 
& \cellcolor{red!4.5!white}11.1 
& \cellcolor{red!2.0!white}40.7 
& \cellcolor{green!11!white}20.6 
& \cellcolor{red!18.0!white}18.9 
& \cellcolor{green!20.0!white}31.7 \\
\xmark & \cmark 
& \cellcolor{red!0.7!white}57.5 
& \cellcolor{red!0.9!white}70.7 
& \cellcolor{green!5!white}9.2 
& \cellcolor{red!15.5!white}10.8 
& \cellcolor{green!22.0!white}20.5 
& \cellcolor{red!11.5!white}37.5 
& \cellcolor{red!11.0!white}50.1 
& \cellcolor{green!18.0!white}8.2 
& \cellcolor{red!9.0!white}16.3 
& \cellcolor{green!48.0!white}13.3 
& \cellcolor{red!12.5!white}3.3 
& \cellcolor{red!19.0!white}23.3 
& \cellcolor{green!40.0!white}6.5 
& \cellcolor{red!20.5!white}20.7 
& \cellcolor{green!100!white}9.5 \\
\cmark & \cmark 
& \cellcolor{red!0.7!white}57.5 
& \cellcolor{green!11.8!white}74.3 
& \cellcolor{red!2!white}10.6 
& \cellcolor{red!15.5!white}10.5 
& \cellcolor{green!2.0!white}25.7 
& \cellcolor{red!13.5!white}35.7 
& \cellcolor{red!7.0!white}53.9 
& \cellcolor{green!2!white}19.7 
& \cellcolor{red!8.0!white}19.6 
& \cellcolor{green!12.0!white}28.4 
& \cellcolor{red!3.5!white}12.2 
& \cellcolor{red!3.0!white}39.5 
& \cellcolor{green!6.0!white}23.7 
& \cellcolor{red!18.0!white}18.0 
& \cellcolor{green!6.0!white}35.2 \\
\bottomrule[0.5mm]
\end{tabular}
\caption{\textbf{Performance comparison of two early-exit approaches}—Extrinsic (Ext.) and Intrinsic (Int.)—vs. the ReAct baseline across four LLMs in three embodied environments. \colorbox{red!20}{Red} indicates \textbf{negative} impact (e.g., performance drop or progress degradation), while \colorbox{green!50}{Green} shows \textbf{positive} effects (e.g., reduced redundancy). Metrics: SR (Success Rate), PR (Progress Rate), RS (Redundant Steps), PD (Progress Degradation), and Steps (Average Steps).}
\label{tab:mainres_embodied}
\end{table*}

\paragraph{Hyperparameters} For all experiments, we set the temperature to 0.1 and limit each turn’s response to a maximum of 256 tokens.

\paragraph{Device and Platform} All experiments are conducted on two NVIDIA A100 GPUs with 80GB of memory each. We deploy the models using VLLM \citep{kwon2023efficient} for distributed inference and access them through OpenAI-compatible chat completion APIs \citep{achiam2023gpt}. Evaluation is performed using AgentBoard \citep{chang2024agentboard}, measuring both success rate and progress rate.

\section{Main Results}

\begin{table*}[h]
\centering
\small
\begin{tabular}{cccccccccccc}
\toprule[0.5mm]
\multicolumn{2}{c}{\textbf{Setting}} & \multicolumn{5}{c}{\textbf{PDDL}} & \multicolumn{5}{c}{\textbf{Jericho}}\\
\cmidrule(lr){1-2} \cmidrule(lr){3-7} \cmidrule(lr){8-12}
\textbf{Int.} & \textbf{Ext.} & SR$\uparrow$ & PR$\uparrow$ & RS$\downarrow$ & PD$\downarrow$ & Steps$\downarrow$ & SR$\uparrow$ & PR$\uparrow$ & RS$\downarrow$ & PD$\downarrow$ & Steps$\downarrow$ \\\midrule
\multicolumn{11}{c}{\textit{Llama3.1-8B-Instruct}} \\\hdashline[3pt/3pt]\noalign{\vskip 0.5ex}
- & - & \cellcolor{red!0.0!white}11.7 & \cellcolor{red!0.0!white}29.9 & \cellcolor{green!0.0!white}27.4 & \cellcolor{red!0.0!white}- & \cellcolor{green!0.0!white}38.3 & \cellcolor{red!0.0!white}5.0 & \cellcolor{red!0.0!white}27.3 & \cellcolor{green!0.0!white}25.2 & \cellcolor{red!0.0!white}- & \cellcolor{green!0.0!white}36.5 \\
\cmark & \xmark & \cellcolor{red!5.0!white}6.7 & \cellcolor{green!2.4!white}30.5 & \cellcolor{green!16!white}19.7 & \cellcolor{red!6.1!white}6.1 & \cellcolor{green!27.6!white}31.4 & \cellcolor{red!0.0!white}5.0 & \cellcolor{red!0.5!white}26.8 & \cellcolor{green!4!white}23.0 & \cellcolor{red!9.0!white}9.0 & \cellcolor{green!4.8!white}37.7 \\
\xmark & \cmark & \cellcolor{red!10.0!white}1.7 & \cellcolor{red!25.5!white}4.4 & \cellcolor{green!46!white}3.6 & \cellcolor{red!25.9!white}25.9 & \cellcolor{green!100!white}4.0 & \cellcolor{red!5.0!white}0.0 & \cellcolor{red!19.8!white}7.5 & \cellcolor{green!39!white}5.4 & \cellcolor{red!19.8!white}19.8 & \cellcolor{green!100!white}6.8 \\
\cmark & \cmark & \cellcolor{red!3.4!white}8.3 & \cellcolor{green!6.0!white}31.4 & \cellcolor{green!14!white}20.4 & \cellcolor{red!6.1!white}6.1 & \cellcolor{green!24.0!white}32.3 & \cellcolor{green!20.0!white}10.0 & \cellcolor{green}31.8 & \cellcolor{green!4!white}23.2 & \cellcolor{red!10.5!white}10.5 & \cellcolor{green!12.8!white}33.3 \\\midrule

\multicolumn{11}{c}{\textit{Llama3.1-70B-Instruct}} \\\hdashline[3pt/3pt]
\noalign{\vskip 0.5ex}
- & - & 45.0 & 62.2 & 16.2  & -   & 31.1 & 35 & 55.9 & 13.9 & -    & 32.3 \\
\cmark & \xmark & \cellcolor{red!3.3!white}41.7 & \cellcolor{green!10.4!white}64.8 & \cellcolor{green!7.6!white}12.4 & \cellcolor{red!5.8!white}5.8 & \cellcolor{green!11.6!white}28.2 & \cellcolor{red!10.0!white}25 & \cellcolor{red!14.4!white}41.5 & \cellcolor{green!2.8!white}13.2 & \cellcolor{red!23.1!white}23.1 & \cellcolor{green!10.0!white}29.8 \\
\xmark & \cmark & \cellcolor{red!1.7!white}43.3 & \cellcolor{green!5.2!white}63.5 & \cellcolor{green!16.0!white}8.4 & \cellcolor{red!4.9!white}4.9 & \cellcolor{green!32.4!white}23 & \cellcolor{red!15.0!white}20 & \cellcolor{red!17.4!white}38.5 & \cellcolor{green!8!white}10.0 & \cellcolor{red!19.8!white}19.8 & \cellcolor{green!42.4!white}21.7 \\
\cmark & \cmark & \cellcolor{red!6.7!white}38.3 & \cellcolor{red!0.3!white}61.9 & \cellcolor{green!4!white}14.2 & \cellcolor{red!8.1!white}8.1 & \cellcolor{green!5.6!white}29.7 & \cellcolor{red!15.0!white}20 & \cellcolor{red!14.4!white}41.5 & \cellcolor{green!0.8!white}14.3 & \cellcolor{red!21.1!white}21.1 & \cellcolor{green!11.6!white}29.4 \\\midrule

\multicolumn{11}{c}{\textit{Mistral-7B-Instruct}} \\\hdashline[3pt/3pt]
\noalign{\vskip 0.5ex}
- & -       & 0    & 9.7  & 37.5 & -   & 40   & 0  & 11.7 & 35.6 & -    & 38.4 \\
\cmark & \xmark & \cellcolor{green!6.8!white}1.7 & \cellcolor{green!9.6!white}12.1 & \cellcolor{green!20!white}27.4 & \cellcolor{red!5.0!white}5 & \cellcolor{green!38.8!white}30.3 & \cellcolor{red!0.0!white}0 & \cellcolor{red!4.8!white}6.9 & \cellcolor{green!13!white}29.1 & \cellcolor{red!4.8!white}4.8 & \cellcolor{green!32.8!white}30.2 \\
\xmark & \cmark & \cellcolor{green!13.2!white}3.3 & \cellcolor{green!16.4!white}13.8 & \cellcolor{green!40!white}17.8 & \cellcolor{red!4.6!white}4.6 & \cellcolor{green!76.4!white}20.9 & \cellcolor{red!0.0!white}0.0 & \cellcolor{red!2.7!white}9 & \cellcolor{green!22!white}23.9 & \cellcolor{red!6.0!white}6 & \cellcolor{green!49.2!white}26.1 \\
\cmark & \cmark & \cellcolor{green!13.2!white}3.3 & \cellcolor{green!12.8!white}12.9 & \cellcolor{green!11!white}31.7 & \cellcolor{red!4.2!white}4.2 & \cellcolor{green!17.6!white}35.6 & \cellcolor{red!0.0!white}0 & \cellcolor{green!1.2!white}12 & \cellcolor{green!2!white}34.7 & \cellcolor{red!3.5!white}3.5 & \cellcolor{green!7.2!white}36.6 \\\midrule

\multicolumn{11}{c}{\textit{Mistral-24B-Instruct}} \\\hdashline[3pt/3pt]
\noalign{\vskip 0.5ex}
- & -   & 13.3 & 27.4 & 28.3  & -   & 37   & 15 & 43.8 & 25.1 & -    & 37.3 \\
\cmark & \xmark & \cellcolor{red!0.0!white}13.3 & \cellcolor{green!23.6!white}33.3 & \cellcolor{red!1.1!white}25.4 & \cellcolor{red!8.5!white}8.5 & \cellcolor{green!9.2!white}34.7 & \cellcolor{red!5.0!white}10 & \cellcolor{red!10.0!white}33.8 & \cellcolor{green!3!white}23.5 & \cellcolor{red!12.7!white}12.7 & \cellcolor{green!18.4!white}32.7 \\
\xmark & \cmark & \cellcolor{red!3.3!white}10.0 & \cellcolor{red!3.1!white}24.3 & \cellcolor{green!14.8!white}10.4 & \cellcolor{red!7.9!white}7.9 & \cellcolor{green!83.2!white}16.2 & \cellcolor{red!10.0!white}5 & \cellcolor{red!14.3!white}29.5 & \cellcolor{green!22!white}14.0 & \cellcolor{red!18.2!white}18.2 & \cellcolor{green!65.6!white}20.9 \\
\cmark & \cmark & \cellcolor{red!1.6!white}11.7 & \cellcolor{green!18.4!white}32.0 & \cellcolor{red!2.4!white}26.6 & \cellcolor{red!7.8!white}7.8 & \cellcolor{green!3.6!white}36.1 & \cellcolor{red!5.0!white}10 & \cellcolor{red!16.4!white}27.4 & \cellcolor{green!4!white}27.6 & \cellcolor{red!20.5!white}20.5 & \cellcolor{green!1.2!white}37.0 \\
\bottomrule[0.5mm]
\end{tabular}
\caption{\textbf{Performance comparison of two early-exit settings} across four LLMs in game environments. \colorbox{red!20}{Red} indicates negative impact, while \colorbox{green!50}{Green} shows positive effects (e.g., reduced redundancy). Metrics: SR (Success Rate), PR (Progress Rate), RS (Redundant Steps), PD (Progress Degradation), and Steps (Average Steps).}
\label{tab:mainres_game}
\end{table*}

We experiment on 3 embodied environments and 2 gaming environments, and report results in Table~\ref{tab:mainres_embodied} and Table~\ref{tab:mainres_game}, respectively. We can see that:

\paragraph{(i) Early-exit mechanisms significantly reduce redundant steps.}
Across all three embodied environments, baseline methods exhibit substantial redundancy (“RS”) in their thought-action sequences. For example, \textit{LLama3.1-8B-Instruct} averages 26.4 unnecessary steps out of 40 in Alfworld\footnote{Note that our implementation of RS did not account for some trivial cases in the first arXiv version. We have corrected this in the current version, and the correction does not affect our main conclusions.}. Almost all early exit mechanisms are able to reduce the redundancy, by approximately 50\% to 70\%, leading to a notable increase in overall efficiency. A similar trend is observed in the average steps (“Steps”), decreasing alongside the reduction in redundant steps, further highlighting the effectiveness of the early-exit mechanism in improving task efficiency.

\paragraph{(ii) Minor performance drop in success and progress rates.}
While early exit improves efficiency, it inevitably causes slight reductions in both success and progress rates. The observed progress degradation (“PD”) further confirms this trade-off. However, for all four tested LLMs, certain early exit strategies yield minimal performance loss. For example, using the extrinsic ("Ext.") method on \textit{Llama3.1-70B-Instruct}, the progress rate drops by under 2\%, 3\%, and 4\% in ALFWorld, BabyAI, and ScienceWorld, respectively. This shows that appropriate early exits can greatly improve efficiency with negligible performance impact.

\paragraph{(iii) LLMs show varying preferences for early exit strategies.}
LLMs respond differently to the same early exit approach. For example, the intrinsic ("Int.") early exit performs better for \textit{Mistral-7B-Instruct}, whereas it significantly degrades the performance of \textit{Mistral-24B-Instruct}. Conversely, \textit{Mistral-24B-Instruct} benefits more from the extrinsic method ("Ext."). This is possibly because the larger Mistral LLMs is more sensitive to intrinsic cues, resulting in premature termination, whereas extrinsic method provide more stable exit signals.

\paragraph{(iv) Combining intrinsic and extrinsic early exit maximizes performance retention.}
We explore a hybrid strategy that first applies extrinsic verification to detect potential exit, then applies the intrinsic method to confirm termination. While this increases the number of steps and reduces efficiency, it achieves the best performance preservation ("Int.\cmark + Ext.\cmark"). Notably, it even slightly improves performance on \textit{Llama3.1-70B-Instruct} and \textit{Mistral-24B-Instruct}, possibly due to diversity behavior introduced by prompt modification.

\paragraph{(v) Early-exit strategy generalizes to gaming environments.} 
As shown in Table~\ref{tab:mainres_game}, applying early-exit in gaming environments yields similar trends, but smaller gains in efficiency and minor performance changes compared to embodied tasks. Redundancy reduction is less pronounced (generally below 50\%), and the drops in performance are marginal, except for \textit{Mistral-7B-Instruct}, occasionally showing improvement. This may be due to: 1) the longer trajectories in gaming environments, which lead to lower baseline success rates (e.g., below 20\% for most LLMs except \textit{Llama3.1-70B-Instruct}) and greater sensitivity to prompt variations; and 2) ambiguous subgoal definitions, allowing multiple valid strategies and reducing consistency in progress measurement.

\section{Analysis}

\subsection{Interpretation of Efficiency Metrics} 

\begin{figure}[h]
\centering
\includegraphics[scale=0.61]{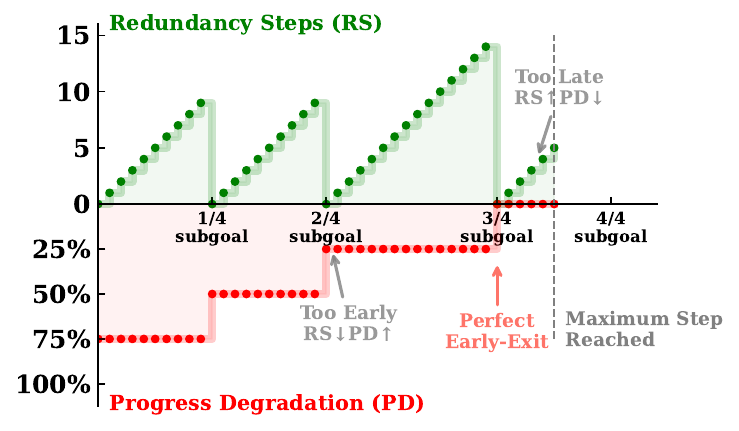}
\caption{\textbf{Redundant Steps and Progress Degradation} measured in a failure case with 3 out of 4 subgoals completed. The metrics vary as the early-exit mechanism is triggered at different steps.}
\label{fig:metric}
\end{figure}

We illustrate how Redundant Steps (RS) and Progress Rate (PR) complement each other in measuring the early-exit behavior in Figure~\ref{fig:metric}.

\paragraph{Perfect Early-Exit Scenario} The ideal early exit scenario ("Perfect Early Exit") occurs when both RS and PR are zero, meaning no redundant steps and no progress loss. However, this ideal is rarely achievable across all environments in practice.

\paragraph{Too-Early Scenarios} If the early exit mechanism triggers too early ("Too Early"), it may reduce redundant steps but significantly impair progress. This is evident in the result of the external early exit of \textit{Llama3.1-8B-Instruct} on BabyAI, where early termination yields a low RS but a high PD of 30.5.

\paragraph{Too-Late Scenarios} Conversely, if the early exit mechanism triggers too late ("Too Late"), PD remains low but RS stays high. This is seen in \textit{Mistral-24B-Instruct}, when using both intrinsic and extrinsic early exit methods fail to reduce RS.

\paragraph{Visualization of Exit Scenarios} Since Figure~\ref{fig:metric} offers only a descriptive illustration of the efficiency metrics, we provide a detailed analysis of the experimental data in Appendix~\ref{appendix:analysis_metrics}, including further insights and interpretations.

\paragraph{Takeaways} Neither too-early nor too-late exits are optimal in practice. Our results highlight the importance of selecting appropriate early exit settings for each LLM to balance RS and PR effectively.

\subsection{Inference Cost}

\begin{figure}[t]
\centering
\includegraphics[scale=0.50]{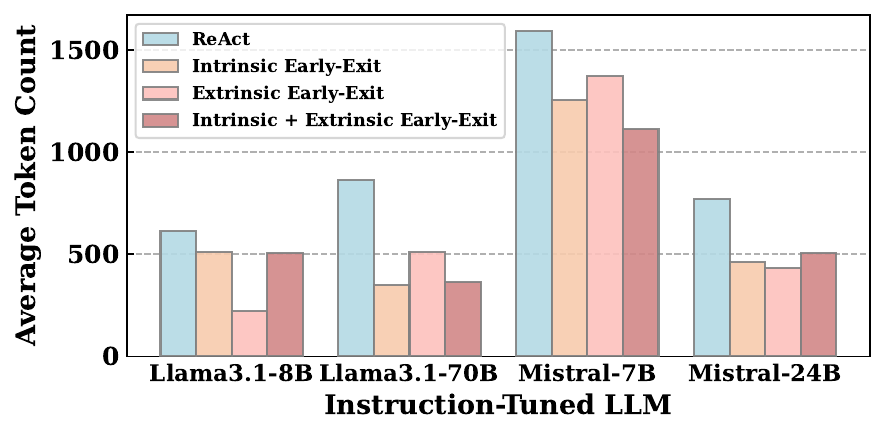}
\caption{\textbf{Comparison of the average token cost} for one environment using different early-exit mechanisms.}
\label{fig:token}
\end{figure}

To further validate the efficiency improvements achieved by the early-exit mechanism, in addition to reporting the average number of execution steps in the main results, we also examine the average token cost for each environment, which directly reflects the computational resource usage. As shown in Figure~\ref{fig:token}, the early-exit approach consistently reduces the number of tokens compared to ReAct across all four tested LLMs. It is worth noting that, in our extrinsic early-exit approach, the verification module generates only a simple "YES" or "NO" response. As a result, it has a negligible impact on the overall token cost.

\section{Practical Implications}

\begin{figure}[t]
\centering
\includegraphics[scale=0.7]{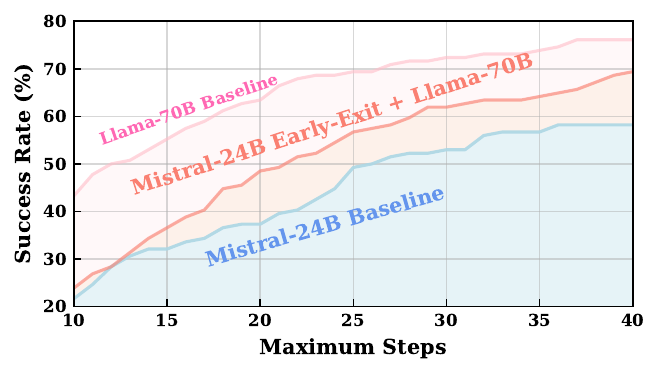}
\caption{\textbf{Performance comparison under different max step limits} using strong-agent assistance with an early-exit weak agent, compared to baseline agents. \textit{Mistral-24B-Instruct} ("Mistral-24B") is used as the weak agent, and \textit{Llama-3.1-70B-Instruct} ("Llama-70B") as the strong agent.} 
\label{fig:practical}
\end{figure}

\begin{figure*}[t]
\centering
\includegraphics[scale=0.44]{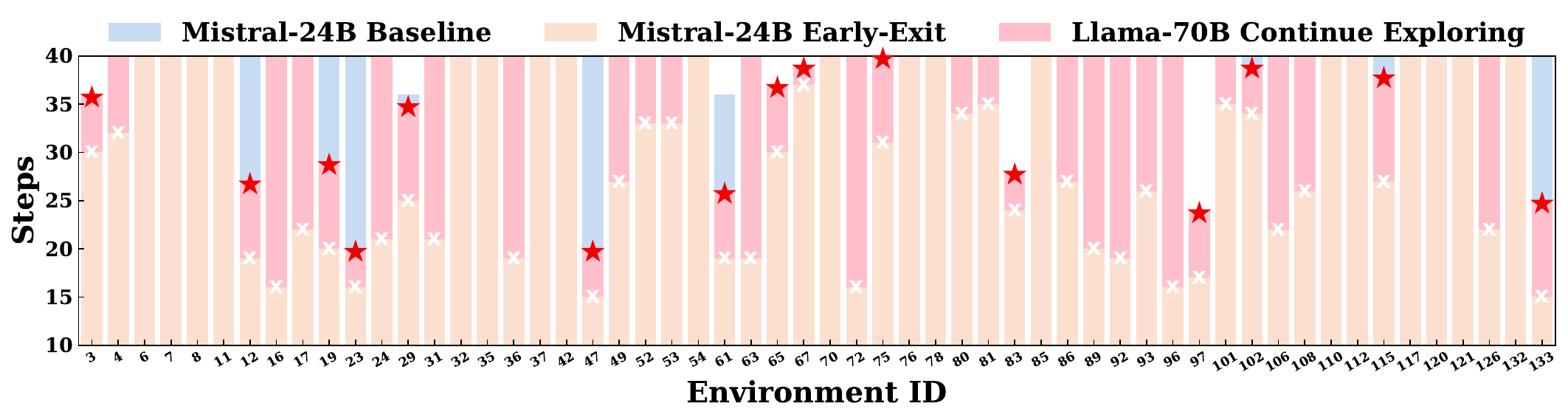}
\caption{\textbf{Case study of failure environments under the early-exit approach} in ALFWorld. Different colors indicate the contributions of various strategies. "x" marks the exit step for each environment, and \textcolor{red}{\ding{72}} indicates completion by the stronger agent (\textit{Llama3.1-70B-Instruct}).}
\label{fig:practical_case}
\end{figure*}

\subsection{Motivation}

A key advantage of our proposed early-exit mechanism is that agents capable of recognizing failure can proactively terminate and seek assistance, thereby increasing the likelihood of successful problem-solving. This aligns with realistic application scenarios, where humans may intervene directly or request help via a central server. In contrast, agents without early-exit continue until the step limit, often wasting valuable interactions and failing to complete the task.

To illustrate this, we simulate a practical scenario in embodied environments, where \textbf{a weaker agent exits early from challenging environments and requests assistance from a stronger agent}.

\subsection{Setting}

\paragraph{Dataset} We use ALFWorld \citep{shridharalfworld} as our test set, which is a typical embodied environment with 134 different tasks.

\paragraph{Models} We use \textit{Mistral-24B-Instruct} as the weak agent which achieves a 58.2\% success rate under a ReAct-style format and 57.5\% when paired with an external early-exit mechanism (seen in Table~\ref{tab:mainres_embodied}), and \textit{Llama3.1-70B-Instruct} as the strong agent.

\paragraph{Setup} In baseline, the weaker agent executes up to 40 steps regardless of progress. With the extrinsic early-exit mechanism, it can terminate early and hand over control to a stronger agent, which replans and continues within the remaining steps. 

\subsection{Experiment}

\paragraph{Result} As shown in Figure~\ref{fig:practical}, early exit followed by strong agent assistance yields over a 10\% improvement in success rate within the same 40-step budget, demonstrating the effectiveness of reallocating interaction steps to a more capable agent.

\paragraph{Token Usage} A potential concern is that introducing a stronger agent does not necessarily improve ``efficiency'' in practice, as larger LLMs typically incur higher inference costs. To quantify this, we report token usage in Table~\ref{tab:practical_token}. The stronger \textit{Llama3.1-70B-Instruct} consumes an average of 128 tokens per instance, whereas the weaker \textit{Mistral-24B-Instruct} requires nearly 100 fewer tokens. This modest increase of computation yields an 11.2\% improvement in success rate, which represents a reasonable trade-off between performance gains and computational overhead.

\begin{table}[h]
\centering
\small
\setlength{\tabcolsep}{4pt}
\begin{tabular}{lccc}
\toprule[0.5mm]
\multirow{2}{*}{\textbf{Setting}} & \multicolumn{2}{c}{\textbf{No. of Tokens}} & \multirow{2}{*}{\textbf{SR (\%)}} \\\cmidrule(lr){2-3}
& Mistral & Llama & \\\midrule
Mistral-24B Base & 622 & - & 58.2 \\
Mistral-24B Early-Exit & 535 & - & 57.5 \\
+ Llama-70B & - & 128 & 69.4 \\
\bottomrule[0.5mm]
\end{tabular}
\caption{\textbf{Comparison of success rates (SR) and average token usage} per environment across different experimental settings.}
\label{tab:practical_token}
\end{table}

\paragraph{Case Study} Figure~\ref{fig:practical_case} visualizes environments impacted by early-exit, where successful environments by both early-exit and baseline are ignored. Around 15 environments (e.g., \#3, \#12) were completed with strong-agent help. Of these, 7 environments (e.g., \#12, \#19) were not completed by the baseline within 40 steps but were solved with early exit and assistance. Only 2 cases (\#29, \#61) were prematurely exited but solved by baseline. Some tasks (e.g., \#4, \#36) remained unsolved by both agents but benefited from reduced wasted computation. These results clearly highlight the efficiency improvements brought by early-exit, especially when supported by stronger agents.

\paragraph{Adaptation to Reflexion Framework} To illustrate the universality of early-exit behavior, we conduct an additional experiment by integrating the early-exit mechanism with the Reflexion framework \citep{shinn2023reflexion}, where the agent reflects on the early-exit trajectory before resuming execution. This adaptation yields improved results, with further details provided in the Appendix~\ref{appendix:reflexion}.

\section{Discussion}

\paragraph{Selecting Early-Exit Strategies} 
We examine three types of early-exit strategies: intrinsic, extrinsic, and hybrid. The intrinsic approach embeds exit instructions directly into the agent’s prompt; the extrinsic approach relies on a separate verification module (sharing the same LLM backbone) to monitor trajectories; and the hybrid approach combines both, using the extrinsic verifier to signal termination and then injecting intrinsic exit instructions for subsequent steps, ensuring alignment between verifier and agent. Experiments show that different LLMs exhibit distinct behaviors under these strategies, preventing us from recommending a universal best setting. Instead, we provide model-specific recommendations in Appendix~\ref{appendix:recommend}, guided by the principle of minimizing progress degradation while reducing redundant steps.

\paragraph{Overall Trends Across Models} 
Analysis of Table~\ref{tab:mainres_embodied}, Table~\ref{tab:mainres_game}, and Figure~\ref{fig:vis_metrics} indicates that the effectiveness of early-exit varies across LLMs. In practice, larger models such as \textit{Llama3.1-70B-Instruct} generally yield more reliable early exits due to stronger instruction-following capabilities. By contrast, smaller models are more prone to premature termination, which can severely harm progress, and thus should be used with caution.

\paragraph{Measurement of Redundant Step} 
We admit that the definition of redundant steps carries inherent ambiguity. In particular, for the first trial in each environment, the true achievable progress is unknown, making it unclear which prior steps should be considered redundant. Once a subgoal is achieved, earlier steps may retroactively be deemed useful rather than redundant. This limitation may skew measurements in certain environments. We hope that researchers in the community will further investigate this issue and develop more sensitive metrics for distinguishing agent behaviors, such as productive reasoning versus unproductive looping. We also remind users to be aware of these limitations during evaluation of agent performance.

\section{Related Work}

\paragraph{LLM-based Agents}
LLM-based agents are central to many tasks and show strong practical potential. Some approaches, like ETO \citep{song-etal-2024-trial} and AgentFLAN \citep{chen-etal-2024-agent}, improve performance through expert trajectory training, achieving better generalization. Others, such as ReAct \citep{yao2023react}, PreAct \citep{fu-etal-2025-preact}, and StateFlow \citep{wu2024stateflow}, focus on prompt design to enhance chain-of-thought (CoT, \citealp{wei2022chain}) reasoning. SwiftSage \citep{lin2023swiftsage} integrates the behavior cloning and planning capabilities to enhance task completion. While effective, these methods often neglect efficiency, especially in failure cases. Complementary post-hoc strategies—like self-reflection \citep{shinn2023reflexion}, trajectory revision \citep{ouyang-li-2023-autoplan}, and experience extraction \citep{zhao2024expel}—help refine future behavior but only after trials conclude. We propose early-exit approaches that improve efficiency and demonstrate the practical benefits of leveraging stronger agents and post-hoc strategies.

\paragraph{Dynamic Early Exit}
Dynamic early exit is an adaptive inference strategy originally introduced in pre-trained language models to reduce computational cost and latency by skipping certain layers during inference \citep{zhou2020bert, sun-etal-2022-simple}. Recent work extends this concept to LLMs to address the issue of excessively long and unpredictable generations. \citet{yang2025dynamic} applies early exit mechanism to truncate outputs at appropriate reasoning steps, thereby mitigating the “overthinking” problem in LLMs \citep{chen2024not}. \citet{wang2025agentdropout} eliminates redundant agent for better token efficiency in agent-collaboration scenarios. In this work, we apply early exit to LLM-based agents in embodied environments, proposing an efficient and robust method adaptable to various agents, along with metrics to assess performance.

\paragraph{Agent Verification and Evaluation} Traditional benchmarks like AgentBench \citep{liuagentbench} assess overall agent performance using metrics such as reward or success rate. AgentBoard \citep{chang2024agentboard} improves transparency with human-annotated subgoals for process-level evaluation. A growing line of work explores using agents themselves as evaluators, extending ideas from text generation \citep{zheng2023judging, lu-etal-2024-error}, code evaluation \citep{chen-etal-2024-agent}. For example, \citet{panautonomous} explore using agents for self-evaluation and refinement. In this work, we leverage the agent verification module to verify its process in extrinsic early exit approach, and introduce two efficiency metrics to complement existing agent evaluation strategies.

\section{Conclusion}
\label{sec:conclusion}

In this work, We propose a dynamic early-exit framework for LLM-based agents in complex embodied environments, incorporating intrinsic and extrinsic early-exit mechanisms. Both approaches improve efficiency in our experiments. To better evaluate the impact of early exits, we introduce two complementary metrics that capture both its positive and negative effects. Additionally, we design a practical experiment in which a stronger agent assists a weaker one in continuing task execution, leading to enhanced performance. Early-exit mechanism is also universally applicable and can be integrated with other techniques. We hope our approach serves as a first step toward improving the efficiency of LLM-based agents and that our proposed metrics can be readily adopted by future research for evaluating agent efficiency.

\section*{Limitations}

The limitations of our work are as follows:

\begin{itemize}
    \item \textbf{Limited Datasets:} We evaluate only five datasets from embodied and gaming environments. Tasks like web navigation or app execution are excluded, as they often involve simpler, more direct goals, making early-exit less impactful. We leave these for future work.

    \item \textbf{No Training Integration:} While our approaches and metrics are designed to be plug-and-play for all LLM-based agents, we restrict our experiments to models that were not trained with held-in data due to uncertainties about the complexity of datasets.

    \item \textbf{LLM Scope:} We test four open-source LLMs due to budget constraints and to avoid data contamination. Proprietary models like GPT \citep{achiam2023gpt} are not included.

    \item \textbf{Residual Redundancy:} While our approach reduces redundant steps, it does not fully eliminate them, likely due to current LLMs’ limited instruction-following ability. Further improvements are still necessary.

    \item \textbf{Real-World Deployment:} Our evaluation is conducted in simulation rather than real-world environments. While this limits external validity, simulation provides a controlled and necessary first step for analyzing core early-exit behaviors. We also give practical implications to better illustrate the practical use of our approach.
    
\end{itemize}

\section*{Ethics Statement}

We take ethical considerations very seriously, and strictly adhere to the Code of Ethics. All procedures performed in this study are in accordance with the ethical standards. This paper explores early-exit mechanisms for LLM-based agents in embodied environments. Our proposed approaches and metrics, does not include statements that induce the model to generate harmful information. Additionally, the approach focuses solely on determining when to terminate agent execution, thereby reducing potential risks. Both the datasets and models used in this paper are publicly available and have been widely adopted by researchers. We ensure that the findings and conclusions of this paper are reported accurately and objectively. No human participants were involved as evaluators or case studies in this work.

\section*{Acknowledgements}

We thank the anonymous reviewers and the area chair for their insightful comments and valuable suggestions. We have revised the paper and corrected a minor calculation error in the tables identified during the review process. This research is supported by the Fundamental Research Funds for the Central Universities 2242025F20002, the National Science Foundation of China under Grant 61973083, the Shenzhen Science and Technology Program JCYJ20210324121213036. Dr Tao’s research is partially supported by NTU RSR and Start Up Grants.

\bibliography{arxiv_version}

\appendix

\section{Recommended Early-Exit Approaches} \label{appendix:recommend}

Based on our experimental results and analysis, we provide a set of recommendations for selecting suitable early-exit approaches for specific LLMs. These guidelines are summarized in Table~\ref{tab:recommendation} and can serve as a reference for future research.

\begin{table}[h]
\centering
\small
\setlength{\tabcolsep}{4pt}
\begin{tabular}{ccc}
\toprule[0.5mm]
\textbf{LLM} & \textbf{Intrinsic} & \textbf{Extrinsic} \\\midrule
Llama3.1-8B-Instruct & \faToggleOn & \faToggleOn \\
Llama3.1-70B-Instruct & \faToggleOn & \faToggleOn \\
Mistral-7B-Instruct & \faToggleOn & \faToggleOff \\
Mistral-24B-Instruct & \faToggleOff & \faToggleOn \\
\bottomrule[0.5mm]
\end{tabular}

\caption{\textbf{Recommendations for selecting early-exit approaches} for different LLMs.}
\label{tab:recommendation}
\end{table}

\section{Prompt Variants} \label{appendix:promptvariant}

\begin{table*}[h]
\centering
\small
\setlength{\tabcolsep}{4pt}
\begin{tabular}{>{\raggedright\arraybackslash}p{3cm}p{6cm}p{6cm}}
\toprule[0.5mm]
\textbf{Early-Exit Approach} & \textbf{Strict Condition} & \textbf{Modest Condition} \\
\midrule
Intrinsic Early-Exit & 
Once the environment appears complete or no further progress is likely, include 'EXIT' in your action to end the task without delay. & 
If you believe the environment is complete, your task is finished, and no further attempts are needed, please include 'EXIT' in your action. \\\noalign{\vskip 0.5ex}\hdashline[3pt/3pt]
\noalign{\vskip 0.5ex}
Prompt$\rightarrow$LLM & \textit{Llama3.1-70B-Instruct}

\textit{Mistral-24B-Instruct} & \textit{Llama3.1-8B-Instruct} 

\textit{Mistral-7B-Instruct} \\\midrule

Extrinsic Early-Exit & 
Evaluate the current history of the agent and determine if it meets any of the following conditions:

1. The recent steps show repetitive actions or the agent appears to be stuck in a loop.

2. The agent repeatedly checks for valid actions but fails to make meaningful progress toward the objective.

3. The agent’s recent thoughts suggest the task is complete and no further steps are necessary.

4. The task is no longer achievable due to high difficulty or significant deviation from the expected course.

If any of the above conditions are met, output “YES”. Otherwise, output “NO” to indicate the agent should continue exploring.
&
Evaluate the agent’s recent history and consider:

1. Whether the agent appears stuck or making little meaningful progress despite repeated attempts.

2. Whether the task seems complete or no longer feasible to pursue.

If you have good reason to believe further steps are unlikely to help, you may output “YES” to suggest stopping. Otherwise, output “NO” and continue exploring. \\\noalign{\vskip 0.5ex}\hdashline[3pt/3pt]
\noalign{\vskip 0.5ex}
Prompt$\rightarrow$LLM & \textit{Llama3.1-8B-Instruct}

\textit{Mistral-7B-Instruct}

\textit{Mistral-24B-Instruct} & \textit{Llama3.1-70B-Instruct} \\
\bottomrule[0.5mm]
\end{tabular}
\caption{Early-Exit prompt context with different condition. We also provide their correspondding LLM used in our approach.}
\label{tab:exitprompt}
\end{table*}

In our initial experiments, we observed that prompts behave differently across various LLMs. For instance, in the case of extrinsic early-exit, \textit{Llama3.1-70B-Instruct} is particularly sensitive—strict prompts can easily trigger an early exit. To address this, we designed two prompt variants for each experimental setting: “Modest Condition” and “Strict Condition.” The Strict Condition uses a firmer tone and outlines more detailed exit criteria, while the Modest Condition is more lenient. We provide the full prompt contexts in Table~\ref{tab:exitprompt}, along with their corresponding compatible LLMs.

\section{Analysis of Early-Exit Mechanism with Efficiency Metrics} \label{appendix:analysis_metrics}

We present a point-wise visualization of the proposed early-exit mechanism using our efficiency metrics: \textit{Redundant Steps} (RS) and \textit{Progress Degradation} (PD). We implement the recommended early-exit strategy, as proposed in Appendix~\ref{appendix:recommend}. Figure~\ref{fig:vis_metrics} highlights some dataset-specific and model-specific behaviors:

\paragraph{ALFWorld} Larger models such as \textit{Llama3.1-70B-Instruct} and \textit{Mistral-24B-Instruct} tend to terminate after more than 15 steps, whereas smaller models often terminate within the first 10 steps.  

\paragraph{BabyAI} The LLMs exhibit distinct patterns. \textit{Llama3.1-8B-Instruct} is relatively conservative, terminating only a few trajectories. \textit{Mistral-7B-Instruct} frequently terminates within the first 10 steps. Among the larger LLMs, \textit{Mistral-24B-Instruct} terminates too early, causing substantial progress loss, while \textit{Llama3.1-70B-Instruct} shows much smaller degradation (generally < 50\%).  

\paragraph{ScienceWorld and PDDL} \textit{Mistral-24B-Instruct} again demonstrates premature termination, warranting caution when deploying this agent. By contrast, the other three models achieve smaller progress degradation and effectively reduce redundant steps.  

\paragraph{Jericho} The number of Early-exit samples are fewer than other datasets, largely due to the dataset’s limited size of only 20 samples. Consequently, the number of early-exit cases is the lowest among all datasets.  

\paragraph{Takeaways} The results indicate that the early-exit strategy has potential but remains imperfect: redundant steps persist, and premature termination can still harm task progress. The PD metric provides a clearer view of these negative impacts, complementing RS as an efficiency measure.

\begin{figure*}[t]
\centering
\includegraphics[scale=0.38]{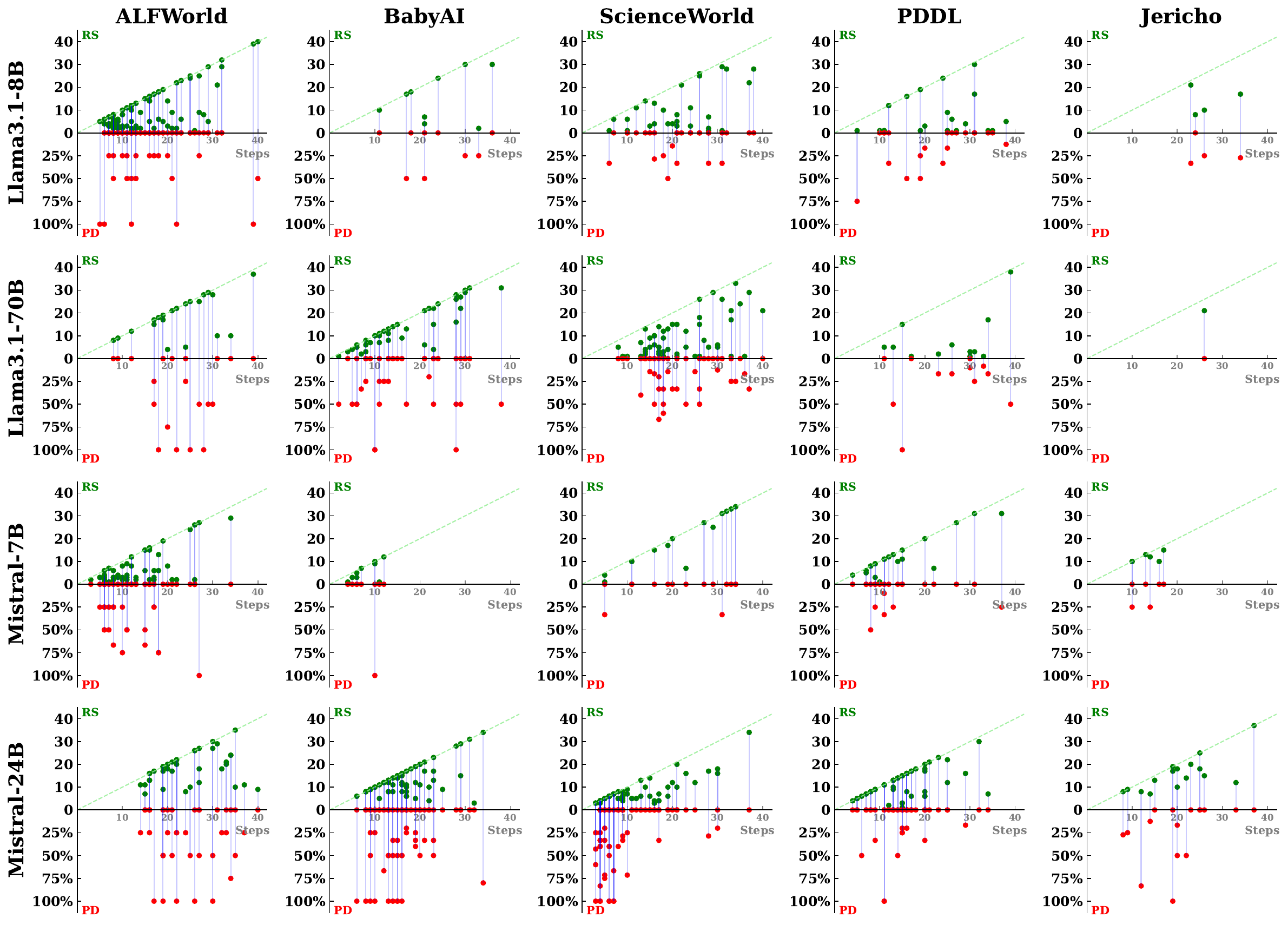}
\caption{Visualization of early-exit effects across four LLMs and multiple datasets. \colorbox{green!50}{Green} and \colorbox{red!50}{red} markers denote redundant steps (RS) and progress degradation (PD), respectively; only early-exit samples are shown. Markers on the \colorbox{green!20}{light-green} dashed line indicates trajectories with no progress before early exit.}
\label{fig:vis_metrics}
\end{figure*}

\section{Adaptation to Reflexion Framework} \label{appendix:reflexion}

To assess the universality of early-exit behavior, we adapt the Reflexion framework for a one-trial scenario: if an agent terminates early, it reflects on its trajectory and resumes the interaction within the same trial. 

\paragraph{Datasets and Models} We evaluate our approach using the \textit{Llama3.1-70B-Instruct} model on the Alfworld dataset.

\paragraph{Settings} To examine early-exit behavior, we employ an external early-exit strategy and integrate the original prompt into the Reflexion framework for self-reflection. In addition to success rate (SR) and progress rate (PR), we also report the average number of steps ("Steps") and the average token usage per environment ("Tokens").

\begin{table}[h]
\centering
\small
\setlength{\tabcolsep}{2pt}
\begin{tabular}{lcccc}
\toprule[0.5mm]
\textbf{Setting} & \textbf{SR (\%)} & \textbf{PR (\%)} & \textbf{Steps} & \textbf{Tokens} \\
\midrule
Baseline (ReAct) & 76.1 & 81.1 & 19.0 & 666 \\
Early-Exit & 70.2 & 79.3 & 13.4 & 459 \\
\textbf{Early-Exit + Reflexion} & \textbf{88.8} & \textbf{92.9} & 18.5 & 661 \\
\bottomrule[0.5mm]
\end{tabular}
\caption{Ablation study of different settings combined with early-exit behavior and the Reflexion framework.}
\label{tab:reflexion}
\end{table}

\paragraph{Results} In Table~\ref{tab:reflexion}, we can see that combining Early Exit with Reflexion significantly boosts both success and progress rates by over 10\%, while keeping the average steps and token costs nearly identical to the baseline. This demonstrates strong compatibility and synergy.

\section{Prompt Context}

We follow \citet{chang2024agentboard} in using the provided task instruction, task goal, and example for each dataset. Since \citet{chang2024agentboard} adopt an act-only prompting style rather than ReAct-style, we follow \citet{song-etal-2024-trial} to design a ReAct-style prompt format. The original examples are extended from Act-Only to ReAct-style using \texttt{gpt-4o-2024-08-06}. Initial observations and interactions are provided by the environment, and the intrinsic and extrinsic early-exit instructions are shown in Table~\ref{tab:exitprompt}. For ALFWorld and ScienceWorld tasks, we observe that providing valid actions leads to a significant performance difference (approximately 10\%–20\% in success rate). Therefore, we include valid actions in these two datasets to ensure fair comparison with prior work \citep{song-etal-2024-trial, fuagentrefine}.

\begin{tcolorbox}[
colback=white!10!white,
colframe=ReActColor,
title=ReAct-Style Prompt for ALFWorld,
breakable]

\textbf{SYSTEM}: 

You are a helpful assistant.

\textbf{USER}: 

Your task is to interact with a virtual household simulator to accomplish a specific task. With each interaction, you will receive an observation. Your role is to ... \hl{\{task instruction\}}

\textbf{ASSISTANT:} OK.

\textbf{USER: }

Here is the example:

\hl{\{example\}}

Now, it's your turn. You should perform thoughts and actions to accomplish the goal. Your response should use the following format:

Thought: <your thoughts>

Action: <your next action>
\\\\
Your task is: \hl{\{task goal\}}

You are in the middle of a room. Looking quickly around you, ... \hl{\{init observation\}}

\hl{\{interaction history\}}
\\\\
\#\# Important \#\#: Your thought should be short, clear and concise. 

\hl{\{intrinsic early-exit instruction\}}
\\\\
The next action could be chosen from these valid actions: \hl{\{valid actions\}}

\end{tcolorbox}

\begin{tcolorbox}[
colback=white!10!white,
colframe=ExitColor,
title=Extrinsic Early-Exit Verification,
breakable]

\textbf{SYSTEM}: 

You are a helpful assistant.

\textbf{USER}: 

You will be given a historical scenario in which you are placed in a specific environment with a designated objective to accomplish.  

\#\#\# Task Description: 
Your task is to interact with a virtual household simulator to accomplish a specific task. With each interaction, you will receive an observation. Your role is to ... \hl{\{task instruction\}}
  
\#\#\# Your Objective: 

\hl{\{task goal\}} 

Your Current History:

\hl{\{interaction history\}}

Instructions:

\hl{\{extrinsic early-exit instruction\}}

Do not include any additional text or explanations in your response.

\end{tcolorbox}

\end{document}